\def\year{2022}\relax
\documentclass[letterpaper]{article} 
\usepackage{aaai22}  
\usepackage{times}  
\usepackage{helvet}  
\usepackage{courier}  
\usepackage[hyphens]{url}  
\usepackage{graphicx} 
\usepackage{natbib}  
\usepackage{caption} 
\DeclareCaptionStyle{ruled}{labelfont=normalfont,labelsep=colon,strut=off} 
\usepackage{algorithm}
\usepackage{algorithmic}

\usepackage{amsmath} 
\usepackage{multirow}
\usepackage{caption} 
\usepackage{wrapfig}
\usepackage{xcolor}
\usepackage{amssymb}
\usepackage{cleveref}
\usepackage{tabularx}
\usepackage{makecell}
\usepackage{booktabs}

\usepackage{newfloat}
\usepackage{listings}
\floatstyle{ruled}
\newfloat{listing}{tb}{lst}{}
\floatname{listing}{Listing}
\title{Label Hallucination for Few-Shot Classification}
\author {
    Yiren Jian,
    Lorenzo Torresani
}
\affiliations {
    Dartmouth College\\
    yiren.jian.gr@dartmouth.edu, LT@dartmouth.edu
}

\usepackage{bibentry}

\begin{document}

\maketitle

\begin{abstract}
Few-shot classification requires adapting knowledge learned from a large annotated base dataset to recognize novel unseen classes, each represented by few labeled examples. In such a scenario, pretraining a network with high capacity on the large dataset and then finetuning it on the few examples causes severe overfitting. At the same time, training a simple linear classifier on top of ``frozen'' features learned from the large labeled dataset fails to adapt the model to the properties of the novel classes, effectively inducing underfitting. In this paper we propose an alternative approach to both of these two popular strategies. First, our method pseudo-labels the entire large dataset using the linear classifier trained on the novel classes. This effectively ``hallucinates'' the novel classes in the large dataset, despite the novel categories not being present in the base database (novel and base classes are disjoint). Then, it finetunes the entire model with a distillation loss on the pseudo-labeled base examples, in addition to the standard cross-entropy loss on the novel dataset. This step effectively trains the network to recognize contextual and appearance cues that are useful for the novel-category recognition but using the entire large-scale base dataset and thus overcoming the inherent data-scarcity problem of few-shot learning. Despite the simplicity of the approach, we show that that our method outperforms the state-of-the-art on four well-established few-shot classification benchmarks.
\end{abstract}

\section{Introduction}
Deep learning has emerged as the prominent learning paradigm for large data scenarios and it has achieved impressive results in wide range of application domains, including computer vision~\cite{AlexNet}, NLP~\cite{devlin-etal-2019-bert} and bioinformatics~\cite{AlphaFold}. However, it remains difficult to adapt deep learning models to settings where few labeled examples are available, since large-capacity models are inherently prone to overfitting. 

Few-shot learning is usually studied under the episodic learning paradigm, which simulates the few-shot setting during training by repeatedly sampling few examples from a small subset of categories of  a large base dataset. Meta-learning algorithms~\cite{MAML, Ravi2017OptimizationAA, Koch2015SiameseNN, MatchNet, ProtoNet} optimized on these training episodes have advanced the  field of few-shot classification. However, recent works~\cite{chen2018a, Dhillon2020A, tian2020rethink} have shown that a pure transfer learning strategy is often more competitive. For example, Tian et al.~\cite{tian2020rethink} proposed to first pretrain a large capacity classification model on the base dataset and then to simply learn a linear classifier on this pretrained representation using the few novel examples. The few-shot performance of the transferred model can be further improved by multiple distillation iterations~\cite{pmlr-v80-furlanello18a}, or by combining several losses simultaneously, e.g., entropy maximization, rotational self-supervision, and knowledge distillation~\cite{rajasegaran2020selfsupervised}.

\begin{figure}[t!]
	\centering
	\includegraphics[width=1.0\linewidth]{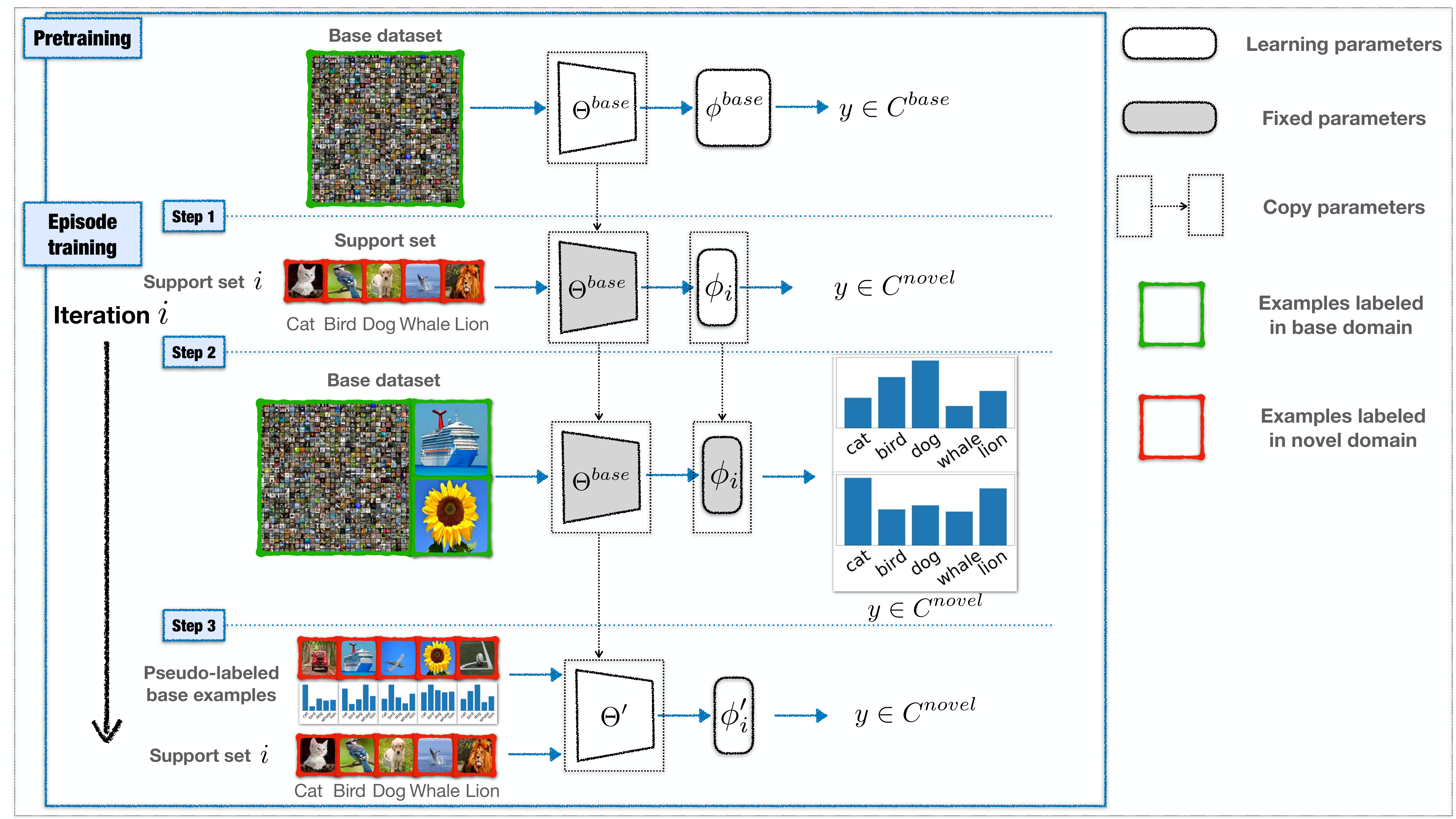}
	\caption{Overview of our proposed approach in an illustrative setting involving 1-shot classification of 5 novel classes. \textit{Pretraining} learns the backbone model $\Theta$ and a classification head $\phi_{0}$ from a labeled base dataset. The backbone is used to compute embeddings for the subsequent stages, while the classification head is discarded. During \textit{Episode training}, step 1) learns a linear classifier $\phi_{1}$ in the novel domain using the support set and the fixed embedding $\Theta$. Step 2) pseudo-labels the base dataset with respect to the label space of the novel domain using the fixed embedding $\Theta$ and the classifier $\phi_{1}$. Step 3) re-learns both the embedding and the classifier with the support set and the pseudo-labeled base dataset using a combination of distillation and cross-entropy maximization. Note that the base dataset and the support set do not share any classes.}
	\label{figure:overview}
\end{figure}

In this paper, we follow the transfer learning approach. However, instead of freezing the representation to the features learned from the base classes~\cite{tian2020rethink, rajasegaran2020selfsupervised, rizve2021exploring}, we finetune the entire model. Since finetuning the network using only the few examples would result in severe overfitting (as evidenced by our ablations), we propose to optimize the model by re-using the entire base dataset but only after having swapped the original labels with pseudo-labels corresponding to the novel classes. This is achieved by running on the base dataset a simple linear classifier trained on the few examples of the novel categories. The classifier effectively ``hallucinates'' the presence of the novel classes in the base images. Although we empirically evaluate our approach in scenarios where the classes of the base dataset are completely disjoint from the novel categories, we demonstrate that this large-scale pseudo-labeled data enables effective finetuning of the entire model for recognition of the novel classes. The optimization is carried out using a combination of distillation over the pseudo-labeled base dataset and cross-entropy minimization over the few-shot examples. An overview of our proposed approach is provided in Fig.~\ref{figure:overview}.

The intuition is that although the novel classes are not ``properly'' represented in the base images, many base examples may include objects that resemble those of the novel classes as encoded by the soft pseudo-labels that define the probabilities of belonging to the novel classes. For example, the pseudo-labeling may assign a probability of 0.6 for a base image of a tiger to belong to the novel class ``domestic cat'' given their appearance similarities. Or it may assign large novel-class pseudo-label probability to a base images because its true base category shares similar contextual background with the novel class, such in the case of ``cars'' and ``pedestrians'' which are both likely to appear in street scenes.  Fine-tuning the entire model on these soft pseudo-labels using a distillation objective (combined with the cross-entropy loss on the few novel image examples) trains the network to recognize these similar or contextual cues on the base dataset, thus steering the representation towards features that are useful for the recognition of the novel classes. Furthermore, because the base dataset is large-scale, these examples serve the role of massive non-parametric data augmentation yielding a representation that is quite general and does not overfit, thus overcoming the data scarcity problem inherent in few-shot learning. We invite the reader to review the visualizations and the explanation in section Visualizations of Label Hallucination of our Technical Appendix (TechApp) for further insights into the behavior of our system. These visualizations confirm the intuition behind our approach , i.e., the fact that the base images with highest scores tend to be those that contain {\em contextual elements} that co-occur with the novel-class objects. Examples include foreground objects that have similar appearance to the few-shot images (e.g., the malamute image in Figure 1 of TechApp), or base examples including objects with shape akin to that of the novel class (e.g., the green mamba in Figure 2 of TechApp, which resembles the shape of a nematode), or even examples matching in terms of spatial layout  (e.g., the images of tobacco shops and upright pianos have similar spatial layout as the bookshop class in Figure 3 of TechApp).

We note that pseudo-labeling has been widely used before for semi-supervised learning where the unlabeled examples belong to the same classes as the labeled ones~\cite{fixmatch, simclrv2, pham2021meta}. Pseudo-labeling has also been adapted to the few-shot setting~\cite{lazarou2020iterative, Wang_2020_CVPR} but still under the empirical setting where novel classes are contained in the unlabeled dataset. The novelty of our work lies in showing that the advantages of pseudo-labeling extend even to the extreme setting where the set of base classes and the set of novel classes are completely disjoint. We also note that our work differs from transductive few-shot learning~\cite{Wang_2020_CVPR, Dhillon2020A} which requires the testing set of unlabeled examples used during the training. Instead, our method operates in a pure inductive setting where within each episode only the small set of novel labeled examples and the base dataset are used for finetuning. 

\section{Related Work}
\textbf{Meta-learning} or learning to learn is the most common approach for few-shot learning. Meta-learning splits the learning into two phases, i.e., a meta-training phase and a meta-testing phase. During each phase, the meta-training set or meta-testing set is organized into multiple episodes, with each episode sampled from the task distribution. Each episode is further partitioned into a small training set and a testing set. In few-shot classification, the training set within each episode has $N$ classes and $K$ examples per class. Meta-learning methods can be further categorized into metric-based versus optimization-based. Metric-based methods~\cite{Koch2015SiameseNN, MatchNet, Sung_2018_CVPR, ProtoNet} learn embeddings for clustering or comparing examples. Few-shot metric learning methods have also been successfully applied to local descriptors \cite{li2019DN4,HUANG2021107935,ZHU2021107797}. Optimization-based methods~\cite{MAML, Flennerhag2020Meta-Learning, li2017metasgd, rusu2018metalearning} learn the parameters of the model or the optimizer for fast adaptation using gradient descent. 

Built upon those classical meta-learning methods, DeepEMD~\cite{Zhang_2020_CVPR} learns a new metric using the Earth Mover's Distance. MetaOptNet~\cite{Lee_2019_CVPR} solves a differentiable convex optimization problem to achieve better generalization of linear classifier. MTL~\cite{Sun_2019_CVPR} explores transfer learning in the meta-learning setting. FEAT~\cite{Ye_2020_CVPR} adapts a set-to-set function to learn task-specific and discriminative embeddings. Neg-Cosine~\cite{neg-cosine} replaces the softmax loss with a negative margin loss in metric learning. MELR~\cite{fei2021melr} adopts an attention module with consistency regularization and explicitly models the relationship between different episodes. Instead of learning unstructured metric, COMET~\cite{cao2021concept} proposes to meta-learn along human-interpretable concepts. ConstellationNet~\cite{xu2021attentional} uses relation learning with self-attention to introduce a cell feature clustering algorithm. PseudoShots~\cite{esfandiarpoor:arxiv20} applies a masking module to select useful features from auxiliary labeled data. IEPT~\cite{zhang2021iept} devises self-supervised pretext tasks at instance level and episode level for few-shot classification. 

\textbf{Transductive/semi-supervised few-shot learning} improves the few-shot results by utilizing the information from the query set or extra unlabeled examples for meta-testing episodes. TIM~\cite{NEURIPS2020_196f5641} maximizes transductive information for few-shot learning. TAFSSL~\cite{TAFSSL} searches discriminative feature sub-spaces for few-shot tasks. ICI~\cite{Wang_2020_CVPR} solves another linear regression hypothesis to filter out less trustworthy instances by pseudo-labeling. The method of Lazarou et al.~\cite{lazarou2020iterative} iteratively refines the pseudo-labels on the unlabeled dataset. Our method is an inductive learning approach, which differs from those used in these works. Without having additional data or information on the query set but only with the base dataset and the support set in each episode, inductive learning is the more common formulation of few-shot classification. 

\textbf{Transfer learning} is the de facto approach for many vision tasks~\cite{pmlr-v32-donahue14}, when the labeled examples are scarce. But transfer learning has seen success in few-shot learning only very recently. New baseline methods~\cite{chen2018a, Dhillon2020A} show competitive performances of few-shot classification by pretraining on a base training set followed by finetuning on the support set from each episode. RFS~\cite{tian2020rethink} outperforms all advanced meta-learning methods at its time by learning a fixed embedding model followed by a linear regression. The success of transfer learning methods relies on the high quality of the pretrained feature embeddings. To get more generalized embeddings of examples, SKD~\cite{rajasegaran2020selfsupervised} proposes to incorporate a rotational self-supervised loss in the pretraining stage. Zhou et al.~\cite{Zhou_2020_CVPR} learn to select a subset of base classes for few-shot classification. Invariant and Equivariant Representations (IER)~\cite{rizve2021exploring} explore contrastive learning during the embedding learning. Chen et al.~\cite{Chen_2021_CVPR} have proposed a self-supervised method that pretrains the  embedding without using labels for the base dataset. Our work also belongs to the genre of transfer learning. But instead of focusing on improving the embedding representation, we study how to better adapt the base knowledge to each novel task. 

Under the transfer learning paradigm, Associative Alignment (AssoAlign)~\cite{Afrasiyabi_2020_ECCV} is the closest to our work. It also exploits the base dataset examples (in the form of a selected subset) to enlarge the novel training set. It aligns the novel examples to the closest base examples in feature space by two strategies: 1) a metric loss to minimize the distance between base examples and the centroids of the novel ones, 2) a conditional Wasserstein adversarial alignment loss. Our method is much simpler than AssoAlign~\cite{Afrasiyabi_2020_ECCV}: it does not require complicated losses, selection of base examples or feature alignment of base dataset to novel examples. We show that by simply finetuning the whole network on a pseudo-labeled version of the base dataset, our method achieves stronger results despite using a smaller model (see \cref{table:ImageNet}, \cref{table:CIFAR}). 

\textbf{Pseudo-labeling}~\cite{pseudo-label} or self-training~\cite{wei2021theoretical} labels the unlabeled dataset first with the model itself, and then re-trains the model with both the labeled and the pseudo-labeled dataset. It has shown great success in semi-supervised learning~\cite{MixMatch, Yu_2020_CVPR, fixmatch}, where pseudo-labeling is applied to a large unlabeled dataset with classes overlapping with the labeled set (the unlabeled and labeled dataset have the same or similar distributions). Algorithms are also designed to filter out low-confident pseudo-labeled examples. The latest work~\cite{pham2021meta} which combines gradient-based meta learning with pseudo-labeling achieves a new state-of-the-art result on ImageNet benchmark~\cite{ImageNet}. The same idea is adapted to semi-supervised few-shot learning as well~\cite{NEURIPS2019_bf25356f, Wang_2020_CVPR, lazarou2020iterative}. Our method has two main differences compared to these approaches. First, the unlabeled and labeled data are from completely disjoint domains. We label the base examples into classes that they do not belong to. Second, we do not have any mechanism to filter the low-confident examples. Our ablation shows that using all the base examples with our pseudo-labels yields better performance than finetuning on only the ones having higher classification probability.

\section{Method}
\subsection{Problem statement}
We now formally define the few-shot classification problem considered in this work. We adopt the common setup which assumes the existence of a large scale labeled base dataset used to discriminatively learn a representation useful for the subsequent novel-class recognition. Let $\mathcal{D}^{base}=\{x^{base}_{t}, y^{base}_{t}\}^{N^{base}}_{t=1}$ be the base dataset, with label $y^{base}_{t} \in \mathcal{C}^{base}$. It is assumed that both the number of classes ($|\mathcal{C}^{base}|$) and the number of examples ($N^{base}$) are large in order to enable good representation learning. We denote with $\mathcal{D}^{novel}=\{x^{novel}_{t}, y^{novel}_{t}\}^{N^{novel}}_{t=1}$ the novel dataset, with $y^{novel}_{t} \in \mathcal{C}^{novel}$. The base classes and novel classes are disjoint, i.e., $\mathcal{C}^{base} \cap \mathcal{C}^{novel} = \varnothing $. We assume the training and testing of the few-shot classification model to be organized in episodes. At each episode $i$, the few-shot learner is given a support set $\mathcal{D}^{support}_{i}=\{x^{support}_{i,t}, y^{support}_{i,t}\}^{NK}_{t=1}$ involving $K$ novel classes and $N$ examples per class sampled from $\mathcal{D}^{novel}$ (with $N$ being very small, typically ranging from 1 to 10). The learner is then evaluated on the query set $\mathcal{D}^{query}_{i}=\{x^{query}_{i,t}, y^{query}_{i,t}\}^{N'K}_{t=1}$, which contains examples of the same $K$ classes as those in $\mathcal{D}^{support}_{i}$. Thus, the query/support sets serve as few-shot training/testing sets, respectively. At each episode $i$, the few-shot learner adapts the representation/model learned from the large-scale $\mathcal{D}^{base}$ to recognize the novel classes given the few training examples in $\mathcal{D}^{support}_{i}$.

\subsection{Learning the embedding representation on the base dataset} \label{section:embedding learning}
We first aim at learning from the base dataset an embedding model that will transfer and generalize well to the downstream few-shot problems. We follow the approach of Tian et al.~\cite{tian2020rethink} (denoted as RFS) and train discriminatively a convolutional neural network consisting of a backbone $f_{\Theta}$ and a final classification layer $g_{\phi}$. The parameters $\{\Theta, \phi\}$ are optimized jointly for the $\left|\mathcal{C}^{base}\right|$-way base classification problem using the dataset $\mathcal{D}^{base}$:
\begin{align}
  \Theta^{base}, \phi^{base} &= \text{argmin}_{\Theta, \phi} \mathbb{E}_{\{x,y\} \in \mathcal{D}^{base}} \mathcal{L}_{CE}(g_{\phi}(f_{\Theta}(x)) , y)
\end{align}
where $\mathcal{L}_{CE}$ is the cross-entropy loss. Prior work has shown the quality of the embedding representation encoded by parameters $\Theta^{base}$ can be further improved by knowledge distillation~\cite{tian2020rethink}, rotational self-supervision~\cite{rajasegaran2020selfsupervised} or by enforcing representations equivalent and invariant to sets of image transformations~\cite{rizve2021exploring}.  In the experiements presented in this paper, we follow the embedding learning strategies of SKD~\cite{rajasegaran2020selfsupervised} (using self-supervised distillation) and IER~\cite{rizve2021exploring} (leveraging invariant and equivariant representations). However, note that our approach is independent of the specific method used for embedding learning. 

\subsection{Hallucinating the presence of novel classes in the base dataset} \label{section:pseduo-labeling}

In order to pseudo-label the base dataset according to the novel classes, we first train a classifier on the support set. For each episode $i$ in the meta-learning phase, we learn a linear classifier $\phi_{i}$ on top of the {\em fixed} feature embedding model $\Theta^{base}$ using the few-shot support set $\mathcal{D}^{support}_{i}=\{x^{support}_{i,t}, y^{support}_{i,t}\}^{NK}_{t=1}$.
\begin{align}
  \phi_{i} &= \text{argmin}_{\phi} \mathbb{E}_{\{x,y\} \in \mathcal{D}_{i}^{support}} \mathcal{L}_{CE}(g_{\phi}(f_{\Theta^{base}}(x)) , y)
\end{align}
Note that in previous works~\cite{tian2020rethink, rajasegaran2020selfsupervised, rizve2021exploring}, $\phi_{i}$ is directly evaluated on query set $\mathcal{D}^{query}_{i}$ to produce the final few-shot classification results. Instead here we use the resulting model $g_{\phi_i}(f_{\Theta^{base}}(x))$ to re-label the base dataset according to the ontology of the novel classes in episode $i$. We denote with $\hat{y}^{base}_{i,t}$ the vector of logits (the outputs before the softmax) generated by applying the learned classifier to example $x^{base}_t$, i.e., $\hat{y}^{base}_{i,t} = g_{\phi_i}(f_{\Theta^{base}}(x_t))$ for $t=1, \hdots, N^{base}$. These soft pseudo-labels are used to retrain the full model via knowledge distillation, as discussed next.

\subsection{Finetuning the whole model to recognize novel classes} \label{section:finetuning-again}
We finally finetune the whole model (i.e., the backbone and the classifier) using mini-batches containing an equal proportion of support and base examples. The loss function for the base examples is knowledge distillation~\cite{KD}, while the objective minimized for the support examples is the cross-entropy (CE). In other words, we optimize the parameters of the model on a mixing of the two losses: 
\begin{align}\label{equation:total-loss}
  \Theta'_{i}, \phi'_{i} = \text{argmin}_{\Theta, \phi} & \lambda_{KL}  \mathbb{E}_{\{x,y\} \in \mathcal{D}^{base}} \mathcal{L}_{KL}(g_{\phi}(f_{\Theta}(x)), \hat{y}) +   \nonumber \\
& \lambda_{CE} \mathbb{E}_{\{x,y\} \in \mathcal{D}_{i}^{support}} \mathcal{L}_{CE}(g_{\phi}(f_{\Theta}(x)) , y)
\end{align}
where $\hat{y}$ denotes the hallucinated pseudo-label, $\mathcal{L}_{KL}$ is the KL divergence between the predictions of the model and the pseudo-labels scaled by temperature $T$, and $\{\lambda_{KL}, \lambda_{CE}\}$ are hyper-parameters trading off the importance of the two losses. Since the support set is quite small (in certain settings, each episode includes five novel classes and only one example for each novel class), we use data augmentation to generate multiple views of each support image, so as to obtain enough examples to fill half of the mini-batch. Specifically, we adopt the standard settings used in prior works~\cite{tian2020rethink, rajasegaran2020selfsupervised, rizve2021exploring} and apply random cropping, color jittering and random flipping to generate multiple views.

Finally, the resulting model $g_{\phi'_i}(f_{\Theta'_i}(x))$ is evaluated on the query set $\mathcal{D}^{query}_{i}=\{x^{novel}_{t}, y^{novel}_{t}\}^{N'K}_{t=1}$. The final results are reported by averaging the accuracies of all episodes.

We note that although the operations of pseudo-labeling and finetuning are presented as separate and in sequence, in practice for certain datasets we found more efficient to generate the target pseudo-labels on the fly for the base examples loaded in the mini-batch without having to store them on disk.

\section{Experiments}

\subsection{Datasets}
We evaluate our method on four widely used few-shot recognition benchmarks: miniImageNet~\cite{MatchNet}, tieredImageNet~\cite{ren2018metalearning}, CIFAR-FS~\cite{bertinetto2018metalearning}, and FC100~\cite{TADAM}. 

\subsection{Experimental setup}
\textbf{Network Architecture}. To make fair comparison to recent works~\cite{tian2020rethink, rajasegaran2020selfsupervised, rizve2021exploring}, we adopt the popular ResNet-12~\cite{He_2016_CVPR} as our backbone.

Further descriptions on the four datasets, the architecture of ResNet-12 we used in experiments and {optimization details} on embedding learning and our finetuning can be found in the Technical Appendix.

\begin{table*}[!h]
  \scriptsize
  \centering
  \begin{tabular}{lccccc}
  \toprule
  {} & {} & \multicolumn{2}{c}{{\bf miniImageNet 5-way}} & \multicolumn{2}{c}{{\bf tieredImageNet 5-way}} \\ 
  \cmidrule(lr){3-4} \cmidrule(lr){5-6}
  {\bf model} & {\bf backbone} & {\bf 1-shot} & {\bf 5-shot} & {\bf 1-shot} & {\bf 5-shot}\\ 
  \midrule
  {ProtoNet~\cite{ProtoNet} (NIPS'17)} & {ResNet-12} & {60.37 $\pm$ 0.83}  & {78.02 $\pm$ 0.57}  & {65.65 $\pm$ 0.92}  & {83.40 $\pm$ 0.65} \\
  {TADAM~\cite{TADAM} (NIPS'18)} & {ResNet-12} & {58.50 $\pm$ 0.30}  & {76.70 $\pm$ 0.30}  & {-} & {-} \\
  {TapNet~\cite{pmlr-v97-yoon19a} (ICML'19)} & {ResNet-12} & {61.65 $\pm$ 0.15}  & {76.36 $\pm$ 0.10}  & {63.08 $\pm$ 0.15}  & {80.26 $\pm$ 0.12} \\
  {MetaOptNet~\cite{Lee_2019_CVPR} (CVPR'19)} & {ResNet-12} & {62.64 $\pm$ 0.61}  & {78.63 $\pm$ 0.46}  & {65.99 $\pm$ 0.72}  & {81.56 $\pm$ 0.53} \\
  {MTL~\cite{Sun_2019_CVPR} (CVPR'19)} & {ResNet-12} & {61.20 $\pm$ 1.80}  & {75.50 $\pm$ 0.80}  & {65.62 $\pm$ 1.80}  & {80.61 $\pm$ 0.90} \\
  {Shot-Free~\cite{Ravichandran_2019_ICCV} (ICCV'19)} & {ResNet-12} & {59.04 $\pm$ 0.43}  & {77.64 $\pm$ 0.39}  & {66.87 $\pm$ 0.43}  & {82.64 $\pm$ 0.43} \\
  {DSN-MR~\cite{Simon_2020_CVPR} (CVPR'20)} & {ResNet-12} & {64.60 $\pm$ 0.72}  & {79.51 $\pm$ 0.50}  & {67.39 $\pm$ 0.83}  & {82.85 $\pm$ 0.56} \\
  {DeepEMD~\cite{Zhang_2020_CVPR} (CVPR'20)} & {ResNet-12} & {65.91 $\pm$ 0.82}  & {82.41 $\pm$ 0.56}  & {71.16 $\pm$ 0.87}  & {86.03 $\pm$ 0.58} \\
  {FEAT~\cite{Ye_2020_CVPR} (CVPR'20)} & {ResNet-12} & {66.78 $\pm$ 0.20}  & {82.05 $\pm$ 0.14}  & {70.80 $\pm$ 0.23}  & {84.79 $\pm$ 0.16} \\
  {Neg-Cosine~\cite{neg-cosine} (ECCV'20)} & {ResNet-12} & {63.85 $\pm$ 0.81}  & {81.57 $\pm$ 0.56}  & {-}  & {-} \\
  {RFS-simple~\cite{tian2020rethink} (ECCV'20)} & {ResNet-12} & {62.02 $\pm$ 0.63}  & {79.64 $\pm$ 0.44}  & {69.74 $\pm$ 0.72}  & {84.41 $\pm$ 0.55} \\
  {RFS-distill~\cite{tian2020rethink} (ECCV'20)} & {ResNet-12} & {64.82 $\pm$ 0.82}  & {82.41 $\pm$ 0.43}  & {71.52 $\pm$ 0.69}  & {86.03 $\pm$ 0.49} \\
  {AssoAlign~\cite{Afrasiyabi_2020_ECCV} (ECCV'20)} & {$\text{ResNet-18}^{\dagger}$} & {59.88 $\pm$ 0.67}  & {80.35 $\pm$ 0.73}  & {69.29 $\pm$ 0.56}  & {85.97 $\pm$ 0.49} \\
  {AssoAlign~\cite{Afrasiyabi_2020_ECCV} (ECCV'20)} & {$\text{WRN-28-10}^{\ddagger}$} & {65.92 $\pm$ 0.60}  & {82.85 $\pm$ 0.55}  & {74.40 $\pm$ 0.68}  & {86.61 $\pm$ 0.59} \\
  {SKD-GEN1~\cite{rajasegaran2020selfsupervised} (Arxiv'20)} & {ResNet-12} & {$66.54 \pm 0.97^{\mathsection}$}  & {$83.18 \pm 0.54^{\mathsection}$}  & {$72.35 \pm 1.23^{\mathsection}$}  & {$85.97 \pm 0.63^{\mathsection}$} \\
  {P-Transfer~\cite{DBLP:journals/corr/abs-2102-03983} (AAAI'21)} & {ResNet-12} & {64.21 $\pm$ 0.77}  & {80.38 $\pm$ 0.59}  & {-} & {-} \\
  {InfoPatch~\cite{gao2021contrastive} (AAAI'21)} & {ResNet-12} & {67.67 $\pm$ 0.45}  & {82.44 $\pm$ 0.31}  & {71.51 $\pm$ 0.52} & {85.44 $\pm$ 0.35} \\
  {MELR~\cite{fei2021melr} (ICLR'21)} & {ResNet-12} & {67.40 $\pm$ 0.43}  & {83.40 $\pm$ 0.28}  & {72.14 $\pm$ 0.51}  & {87.01 $\pm$ 0.35} \\
  {IEPT~\cite{zhang2021iept} (ICLR'21)} & {ResNet-12} & {67.05 $\pm$ 0.44}  & {82.90 $\pm$ 0.30}  & {72.24 $\pm$ 0.50}  & {86.73 $\pm$ 0.34} \\
  {IER-distill~\cite{rizve2021exploring} (CVPR'21)} & {ResNet-12} & {$66.85 \pm 0.76^{\mathsection}$} & {$84.50 \pm 0.53^{\mathsection}$} & {$72.74 \pm 1.25^{\mathsection}$} & {$86.57 \pm 0.81^{\mathsection}$}\\
  \midrule
  {Label-Halluc (pretrained w/ {SKD-GEN1})} & {ResNet-12} & {67.50 $\pm$ 1.01}  & {85.60 $\pm$ 0.52}  & {72.80 $\pm$ 1.20} & {86.93 $\pm$ 0.60} \\
  {Label-Halluc (pretrained w/ {IER-distill})} & {ResNet-12} & {\bf 68.28 $\pm$ 0.77} & {\bf 86.54 $\pm$ 0.46} & {\bf 73.34 $\pm$ 1.25} & {\bf 87.68 $\pm$ 0.83}\\
  \bottomrule
  \end{tabular}
  \caption{Comparison of our method (Label-Halluc) against the state-of-the-art on miniImageNet and tieredImageNet. We report our results with 95\% confidence intervals on meta-testing split of miniImageNet and tieredImageNet. Training is done on the training split only. $\dagger$ indicates using a higher resolution of training images. $\ddagger$ indicates a larger model than ResNet-12. $\mathsection$ indicates our implementations. This makes the fairest comparisons to ours by allowing that those methods are evaluated on exact same episodes.}
  \label{table:ImageNet}
 \end{table*}
 
\subsection{Results on ImageNet-based few-shot benchmarks}\label{section:results-ImageNet}

Table~\ref{table:ImageNet} provides a comparison between our approach and the state-of-the-art in few-show classification on the two ImageNet-based few-shot benchmarks. Our method is denoted as Label-Halluc. On \textbf{miniImageNet}, our method using the SKD pretraining of the backbone yields an absolute improvement of  0.96\% over SKD-GEN1 in the one-shot setting. The improvement become more substantial under the 5-shot setting, with our method producing a gain of 2.42\% over SKD-GEN1. When pretrained with IER~\cite{rizve2021exploring}, our approach achieves one-shot classification accuracy of $68.28 \pm 0.77$, which is over 1.4\% better than all reported results. Under the 5-shot setting, our method improves by 2.04\% over IER-distill which had the best reported number, yielding a new state-of-the-art accuracy of 86.54\%. On the \textbf{tieredImageNet} benchmark, our method pretrained with SKD performs on par with concurrent works~\cite{fei2021melr, zhang2021iept} and outperforms SKD~\cite{rajasegaran2020selfsupervised} by 0.45\% under the 1-shot setting and by 0.96\% under the 5-shot setting. When pretrained with IER, our approach improves over IER-distill by 0.60\% and 1.11\% under the 1-shot and 5-shot settings, respectively, yielding a new state-of-the-art even for this benchmark. 
  
\begin{table*}[!h]
  \scriptsize
  \centering
  \begin{tabular}{lccccc}
  \toprule
  {} & {} & \multicolumn{2}{c}{{\bf CIFAR-FS 5-way}} & \multicolumn{2}{c}{{\bf FC-100 5-way}} \\ 
  \cmidrule(lr){3-4} \cmidrule(lr){5-6}
  {\bf model} & {\bf backbone} & {\bf 1-shot} & {\bf 5-shot} & {\bf 1-shot} & {\bf 5-shot}\\ 
  \midrule
  {ProtoNet~\cite{ProtoNet} (NIPS'17)} & {ResNet-12} & {72.2 $\pm$ 0.7}  & {83.5 $\pm$ 0.5}  & {37.5 $\pm$ 0.6}  & {52.5 $\pm$ 0.6} \\
  {TADAM~\cite{TADAM} (NIPS'18)} & {ResNet-12}  & {-} & {-} & {40.1 $\pm$ 0.4}  & {56.1 $\pm$ 0.4} \\
  {MetaOptNet~\cite{Lee_2019_CVPR} (CVPR'19)} & {ResNet-12} & {72.6 $\pm$ 0.7}  & {84.3 $\pm$ 0.5}  & {41.1 $\pm$ 0.6}  & {55.5 $\pm$ 0.6} \\
  {MTL~\cite{Sun_2019_CVPR} (CVPR'19)} & {ResNet-12} & {-}  & {-}  & {45.1 $\pm$ 1.8}  & {57.6 $\pm$ 0.9} \\
  {Shot-Free~\cite{Ravichandran_2019_ICCV} (ICCV'19)} & {ResNet-12} & {69.2 $\pm$ n/a}  & {84.7 $\pm$ n/a}  & {-}  & {-} \\
  {DSN-MR~\cite{Simon_2020_CVPR} (CVPR'20)} & {ResNet-12} & {75.6 $\pm$ 0.9}  & {86.2 $\pm$ 0.6}  & {-}  & {-} \\
  {DeepEMD~\cite{Zhang_2020_CVPR} (CVPR'20)} & {ResNet-12} & {-}  & {-}  & {46.5 $\pm$ 0.8}  & {63.2 $\pm$ 0.7} \\
  {RFS-simple~\cite{tian2020rethink} (ECCV'20)} & {ResNet-12} & {71.5 $\pm$ 0.8}  & {86.0 $\pm$ 0.5}  & {42.6 $\pm$ 0.7}  & {59.1 $\pm$ 0.6} \\
  {RFS-distill~\cite{tian2020rethink} (ECCV'20)} & {ResNet-12} & {73.9 $\pm$ 0.8}  & {86.9 $\pm$ 0.5}  & {44.6 $\pm$ 0.7}  & {60.9 $\pm$ 0.6} \\
  {AssoAlign~\cite{Afrasiyabi_2020_ECCV} (ECCV'20)} & {$\text{ResNet-18}^{\ddagger}$ } & {-}  & {-}  & {45.8 $\pm$ 0.5}  & {59.7 $\pm$ 0.6} \\
  {SKD-GEN1~\cite{rajasegaran2020selfsupervised} (Arxiv'20)} & {ResNet-12} & {$76.6 \pm 0.9^{\mathsection}$}  & {$88.6 \pm 0.5^{\mathsection}$}  & {$46.5 \pm 0.8^{\mathsection}$}  & {$64.2 \pm 0.8^{\mathsection}$} \\
  {InfoPatch~\cite{gao2021contrastive} (AAAI'21)} & {ResNet-12} & {-}  & {-}  & {43.8 $\pm$ 0.4} & {58.0 $\pm$ 0.4} \\
  {IER-distill~\cite{rizve2021exploring} (CVPR'21)} & {ResNet-12} & {$77.6 \pm 1.0^{\mathsection}$}  & {$89.7 \pm 0.6^{\mathsection}$}  & {$48.1 \pm 0.8^{\mathsection}$} & {$65.0 \pm 0.7^{\mathsection}$} \\
  \midrule
  {Label-Halluc (pretrained w/ {SKD-GEN1})} & {ResNet-12} & {77.3 $\pm$ 0.9}  & {89.5 $\pm$ 0.5}  & {47.3 $\pm$ 0.8} & { 67.2 $\pm$ 0.8} \\
  {Label-Halluc (pretrained w/ {IER-distill})} & {ResNet-12} & {\bf 78.0 $\pm$ 1.0}  & {\bf 90.5 $\pm$ 0.6}  & {\bf 49.1 $\pm$ 0.8} & {\bf 68.0 $\pm$ 0.7}\\
  \bottomrule
  \end{tabular}
  \caption{Comparison of Label-Halluc (ours) to prior works on CIFAR-FS and FC-100. We report our results with 95\% confidence intervals on meta-testing split of CIFAR-FS and FC-100. Training is done on the training split only. $\ddagger$ indicates a different model.  $\mathsection$ indicates our implementations.}
  \label{table:CIFAR}
 \end{table*}
 
\subsection{Results on CIFAR-based few-shot benchmarks}\label{section:results-CIFAR}

Table~\ref{table:CIFAR} compares our method, Label-Halluc, against the state-of-the-art on the two CIFAR-based few-shot benchmarks. On \textbf{CIFAR-FS}, the improvements over SKD-GEN1 (our implementation) for 1-shot and 5-shot are 0.7\% and 0.9\%, respectively. Note that these gains derive exclusively from the addition of the distillation over pseudo-labeled base examples. When using IER-distill as embedding learning, our method improves the baseline by 0.4\% and 0.8\% in the 1-shot and the 5-shot settings, respectively. On \textbf{FC100}, our method achieves improves over the best reported numbers by 0.8\% and 3.0\% in the 1-shot and 5-shot setting, respectively, when pretrained with SKD. The improvements are 1.0\% and 3.0\% when pretrained with IER.

\subsection{Ablations}

The ablation studies are performed to validate improved transfer performances via the use of pseudo-labeled base examples, the use of distillation loss on soft-labels over one-hot encoding labels and finetuning the whole network over learning only the classifier. We further study the effect of different embedding learning methods. Finally, we make an apple-to-apple experimental comparison to AssoAlign~\cite{Afrasiyabi_2020_ECCV}, which also exploits the use of the base dataset during finetuning.

Unless otherwise stated, all the experiments in this section (Ablations) use ResNet-12 and the embedding training by SKD-GEN1~\cite{rajasegaran2020selfsupervised}.

\begin{table}[!h]
  \centering
  \scriptsize
  \setlength{\tabcolsep}{1.5pt}
  \begin{tabular}{ccccccc}
  \toprule
  {} & \multicolumn{2}{c}{{\bf mini-IN}} & \multicolumn{2}{c}{{\bf CIFAR-FS}} & \multicolumn{2}{c}{{\bf FC100}}\\ 
  {} &{1-shot} & {5-shot}  & {1-shot} & {5-shot} & {1-shot} & {5-shot}\\ 
  \midrule
  {\bf\tiny Transfer w/ frozen backbone (LR)} & {66.54} & {83.18} &  {76.6} & {88.6} & {46.5} & {64.2} \\
  {\bf\tiny Transfer w/ finetuning} & {61.43} & {80.03}  & {68.8} & {85.7} & {43.1} & {61.9} \\
  {\bf\tiny Hard LabelHalluc + finetuning} & {65.04} & {80.68} & {75.3} & {85.3} & {44.6} & {62.4} \\
  {\bf\tiny Soft LabelHalluc + finetuning} & {\bf 67.50} & {\bf 85.60}  & {\bf 77.3} & {\bf 89.5} & {\bf 47.3} & {\bf 67.2}\ \\
  \bottomrule
  \end{tabular}
  \caption{Ablation study on different strategy to transfer the base knowledge to the novel-class recognition. Our approach in its complete form (\textit{Soft LabelHalluc + finetuning}) achieves gains over traditional transfer learning (\textit{Transfer w/ frozen backbone (LR)} and \textit{Transfer w/ finetuning}) as well as a variant of our label hallucination method using hard pseudo-labeling (\textit{Hard LabelHalluc + finetuning}).}
  \label{table:ablation-labels}
 \end{table}

\begin{table}[!h]
  \scriptsize
  \centering
  \setlength{\tabcolsep}{6pt}
  \begin{tabular}{cccccc}
  \toprule
  \multicolumn{2}{c}{{\bf Support}}& \multicolumn{2}{c}{{\bf Base}}& \multicolumn{2}{c}{{\bf miniImageNet}} \\ 
  {\bf Net}& {\bf Clf} & {\bf Net}& {\bf Clf}  &{1-shot} & {5-shot} \\ 
  \midrule
  {$\checkmark$} & {$\checkmark$} & {} & {} & {61.43} & {80.03} \\
  {$\checkmark$} & {$\checkmark$} & {} & {$\checkmark$} & {63.59} & {81.53}\\
  {$\checkmark$} & {$\checkmark$} & {$\checkmark$} & {} & {66.18} & {84.36} \\
  {$\checkmark$} & {$\checkmark$} & {$\checkmark$} & {$\checkmark$} & {\bf 67.50} & {\bf 85.60} \\
  \bottomrule
  \end{tabular}
  \caption{Ablation study on gradients from base examples. We study which part of the model (the embedding or the linear classifier) benefits from the use of pseudo-labeled base examples. Having no gradient on either \textit{Net} or \textit{Clf} is equal to finetuning with the novel support set only. And Having gradients on both parts is the default setting of our method.}
  \label{table:ablation-net-clf}
\end{table}
 
\subsubsection{Benefits of the pseudo-labeled base dataset}

In Table~\ref{table:ablation-labels} we ablate on different strategies to transfer the base knowledge to the novel-class recognition. \textit{Transfer w/ frozen backbone (LR)} is the traditional transfer learning procedure of using the few-shot examples to train a linear regression model on top of the frozen backbone learned from the base dataset.  \textit{Transfer w/ finetuning} uses the support set {\em only} to finetune the entire model (backbone and classifier). \textit{Hard LabelHalluc + finetuning} is a variant of our approach where the base dataset is pseudo-labeled with one-hot hard novel-class labels and the entire model is subsequently finetuned using cross-entropy over the base and support set. Finally, \textit{Soft LabelHalluc + finetuning} is our method using soft pseudo-labels. Note that we train \textit{Transfer w/ finetuning} and  {\bf Hard LabelHalluc + finetuning} for 300 steps only for both 1-shot and 5-shot, since we find that training longer in these two settings leads to worse results. From the results in this Table we can infer several findings. First of all, we can observe that \textit{Transfer w/ frozen backbone} performs much better than  \textit{Transfer w/ finetuning}. This confirms the findings of previous works~\cite{tian2020rethink, Afrasiyabi_2020_ECCV, chen2018a} which observed that learning a regression model with the fixed embedding leads to better performance over finetuning the whole network~\cite{Dhillon2020A}. This happens because finetuning the entire network only with the support set results in overfitting. With the combination of distillation loss on the soft-labeled base dataset, our method (\textit{Soft LabelHalluc + finetuning}) eliminates the overfitting problem yielding 5-shot gains of 5.57\%, 3.8\% and 5.3\% on  miniImageNet, CIFAR-FS, and FC100, respectively, over finetuning with the episode examples only. It can be observed that our improves also over the simple strategy of \textit{Transfer w/ frozen backbone} since it provides the advantage of adapting the backbone representation to the specific characteristics of the novel classes.

\begin{table}[!t]
  \scriptsize
  \centering
      \begin{tabular}{lcccccc}
      \toprule
      & \multicolumn{2}{c}{{\bf miniImageNet}} & \multicolumn{2}{c}{{\bf CIFAR-FS}} & \multicolumn{2}{c}{{\bf FC100}} \\
      & {LR} & {ours} & {LR} & {ours} & {LR} & {ours} \\ 
      \midrule
      {RFS-simple~\cite{tian2020rethink}} & {79.33} & {\bf 81.75} & {86.6} & {\bf 87.3} & {58.1} & {\bf 61.2} \\
      {RFS-distill~\cite{tian2020rethink}} & {81.15} & {\bf 82.74} & {86.5} & {\bf 87.1} & {61.0} & {\bf 63.9} \\
      {SKD-gen0~\cite{rajasegaran2020selfsupervised}} & {82.31} & {\bf 84.14} & {87.8} & {\bf 88.8} & {62.8} & {\bf 66.5} \\
      {SKD-gen1~\cite{rajasegaran2020selfsupervised}} & {83.18} & {\bf 85.60} & {88.6} & {\bf 89.5} & {64.2} & {\bf 67.2} \\
      {IER-gen0~\cite{rizve2021exploring}} & {83.88} & {\bf 85.86} & {89.5} & {\bf 90.2} & {63.8} &{\bf 67.2}\\
      {IER-distill~\cite{rizve2021exploring}} & {84.50} & {\bf 86.54} & {89.7} & {\bf 90.5} & {65.0} & {\bf 68.0}  \\
      \midrule
      {\tiny{Average improvement}} & {} & {\bf +2.05} & {} & {\bf +0.8} & {} & {\bf +3.2}  \\
      \bottomrule
      \end{tabular}
      \captionof{table}{Ablation study on different embedding learning methods in 5-shot classification. \textit{LR} is Linear Regression with a fixed embedding.}
      \label{table:ablation-pretrain}
\end{table}

\begin{table}[!t]
  
  \scriptsize
  \centering
      \begin{tabular}{ccccccc}
      \toprule
      & & & & {\bf mini} & {\bf FS} & {\bf FC} \\ 
      & {\bf CA} & {\bf Sub}& {\bf KD}  &{5-shot} & {5-shot} & {5-shot} \\ 
      \midrule
      {\it finetune (softMax)} & {} & {} & {} & {80.03} & {85.7} & {61.9}\\
      {\it \tiny{AssoAlign} (arcMax)}  & {} & {$\checkmark$} & {} &  {81.21} & {85.3} &{60.6}\\
      {\it \tiny{AssoAlign} (arcMax)}  & {$\checkmark$} & {$\checkmark$} & {} & {82.38} & {86.3} & {62.6}\\
      {\it \tiny{AssoAlign} (softMax)} & {$\checkmark$} & {$\checkmark$} & {} & {83.47} & {87.6} & {63.6}\\
      {\it Ours (softMax)} & {} & {$\checkmark$} & {$\checkmark$} & {85.18} & {89.3} & {66.8}\\
      {\it Ours (softMax)} & {} & {} & {$\checkmark$} & {\bf 85.60} & {\bf 89.5} & {\bf 67.2}\\
      \bottomrule
      \end{tabular}
      \captionof{table}{We compare our method to AssoAlign on ResNet-12 in miniImageNet (\textit{mini}) with 84$\times$84 image resolution, CIFAR-FS (\textit{FS}) and FC100 (\textit{FC}). \textit{CA} is the centroid alignment introduced by AssoAlign. \textit{Sub} indicates the use of the scoring matrix in AssoAlign for selection of base examples. \textit{KD} is our method which uses distillation loss on pseudo-labeled base dataset.}
      \label{table:ablation-assoalign}
\end{table}

\subsubsection{Soft or hard labels}
The distillation loss on the pseudo-labeled base dataset is the KL-divergence between the predictions of the linear classifier (soft-labels) and the predictions of the current model. We adopt soft-labeling because hard-labeling examples with novel categories that are not truly represented in those images has a negative effect. This can be clearly observed by the poor performance of \textit{Hard LabelHalluc + finetuning} in Table~\ref{table:ablation-labels}. The use of soft labels over hard labels contributes 5-shot classification gains of 4.92\% in miniImageNet, 4.2\%  in CIFAR-FS, and 4.8\% in FC100. Using the distillation loss with soft-labels is crucial for the success of our method. 

\subsubsection{Finetuning the backbone vs the classifier with hallucinated pseudo-labels}

In this section, we aim at studying whether it is beneficial to finetune the whole network with the hallucinated pseudo-labels, as opposed to just the classifier. We expect that training only the linear classifier (which has few parameters) would not reap the full benefits of the extended re-labeling of the large-scale base dataset, whereas unfreezing the high capacity backbone network may yield further gains. Table~\ref{table:ablation-net-clf} shows the comparisons between 1) applying no gradient from hallucinated pseudo-labels, 2) applying gradients from hallucinated pseudo-labels to the final classifier only, 3) applying gradients from hallucinated pseudo-labels to the backbone feature network only and 4) applying gradients from hallucinated pseudo-labels to the whole network, which is our default setting. As we can see, applying gradients from hallucinated pseudo-labels to the final classifier only leads to small grains (2.16\% and 1.50\% in 1-shot and 5-shot respectively). While learning the backbone feature network with gradients from hallucinated pseudo-labels solely already increases by 4.65\% and 4.33\% in 1-shot and 5-shot. Most of the improvements of our method are coming from learning the backbone feature network with pseudo-labeled base examples.

\subsubsection{Different embedding learning methods}
Our finetuning approach can be used with different embedding learning strategies for pretraining the backbone from the base dataset. We experiment with six different pretraining methods proposed in RFS~\cite{tian2020rethink}, SKD~\cite{rajasegaran2020selfsupervised} and IER~\cite{rizve2021exploring}. Table~\ref{table:ablation-pretrain} shows that our approach constantly improves the classification accuracies over the linear regression (LR) with fixed embeddings. In miniImageNet 5-shot classification, our method has an average 2.05\% improvement over \textit{LR}. In CIFAR-FS and FC100 5-shot, the average improvement are 0.8\% and 3.2\% respectively.
 
\subsubsection{Comparing to AssoAlign} \label{section:AssoAlign}
AssoAlign~\cite{Afrasiyabi_2020_ECCV} also exploits examples from the base dataset to extend the finetuning dataset. The key differences to our method are: 1) AssoAlign experiments with both arcMax and softMax cross-entropy loss; 2) AssoAlign uses a similarity matrix for selecting a subset of base dataset (this requires searching another hyper-parameter to control how many examples to select), while our method uses the entire base dataset; 3) AssoAlign uses centroid alignments on feature space between novel examples and the selected base examples, while our approach uses a distillation loss on the pseudo-labeled base dataset. 

The original AssoAlign is implemented with a different optimizer (Adam) and a different image resolution (224$\times$224 in miniImageNet with ResNet-18). Its backbones (Conv4$\times$4, RestNet-18 or WRN-28-10) differ from those used in recent works (84$\times$84 resolution with ResNet-12). The published results of AssoAlign can be found in \cref{table:ImageNet} and \cref{table:CIFAR}. However, in order to assess AssoAlign and our method on equal ground, we apply the public implementation of AssoAlign to our setting. Starting from the same data augmentations and the same learning policy of ours with their suggested hyper-parameters for ResNet. We report results for AssoAlign finetuned for 100 steps with SGD, since we found this to be the optimal number of steps and the optimizer for this method (we tried both SGD  and Adam with 50, 100, 150, 200 250, 300, 350 and 400 steps). Both our method and AssoAlign here use the same embedding model pretrained with SKD-GEN1~\cite{rajasegaran2020selfsupervised}.

As shown in \cref{table:ablation-assoalign}, AssoAlign alleviates the overfitting issue of finetuning on the support set only (first row) for all three benchmarks. But our method achieves superior results over AssoAlign in all three datasets by only using a simple distillation loss with pseudo-labeled base examples. Though using the arcMax alone yieds an improvement of 1.18\% over \textit{finetune} in miniImageNet, we find that combining arcMax with centroid alignment leads to inferior results in our experimental setup based on ResNet-12 and SGD. AssoAlign with softMax and centroid alignment outperforms \textit{finetune} by 3.44\%, 1.9\% and 2.7\%, whereas our method outperforms AssoAlign by 2.13\%, 1.9\% and 3.6\% in miniImageNet, CIFAR-FS and FC100 respectively.  

\subsubsection{Computational Cost} \label{section:CompCost}
Finetuning methods have a high latency, due to the fact that it requires optimizing all parameters in a large deep neural network for each episode. This is true for our method, AssoAlign and prior works~\cite{Dhillon2020A, liu2018learning}. However, due to the simplicity of our method, the average running time for each training step is 0.35 second only (a total of 105 seconds for each episode), compared to AssoAlign's 0.87 second (87 seconds for each episode) in miniImageNet 5-shot. Furthermore, we invite the reader to review the section "Speeding up the training" in the Technical Appendix, where we present and quantitatively evaluate strategies to lower the total training cost of our approach by more than 66\% while still maintaining significantly higher accuracy compared to the state-of-the-art.

\section{Additional experiments}

We refer the reader to the Technical Appendix for several additional experiments, including:
\begin{itemize}
\item Several visualizations of label hallucination providing useful insights into the effectiveness of our system.
\item Discussion and evaluation of strategies to reduce the computational cost of the training procedure.
\item Performance as a function of the base dataset size.
\item Stochastic finetuning using random mini-batches of the support set.
\item Results for large number of novel classes (10-way and 20-way experiments) in each episode.
\item Results when the base classes and the novel categories are far apart.
\item Experiments with imbalanced base classes.
\item Simultaneous recognition of base and novel classes.
\item Ablation on knowledge distillation.
\item Experiment with feature distillation.
\end{itemize}

\section{Conclusion}
We propose the simple strategy of label hallucination to enable effective finetuning of large-capacity models from few-shot examples of the novel classes. Results on four well-established few-shot classification benchmarks show that even in the extreme scenario where the labels of the base dataset and the labels of the novel examples are completely disjoint, our procedure achieves state-of-the-art accuracy and consistently improves over popular strategies of transfer learning via finetuning or methods that perform linear classification on top of pretrained representations.  

{\small
\bibliographystyle{abbrv}
\bibliography{aaai22}

\begin{thebibliography}{59}
\providecommand{\natexlab}[1]{#1}

\bibitem[{Afrasiyabi, Lalonde, and Gagn\'{e}(2020)}]{Afrasiyabi_2020_ECCV}
Afrasiyabi, A.; Lalonde, J.-F.; and Gagn\'{e}, C. 2020.
\newblock Associative Alignment for Few-shot Image Classification.
\newblock In \emph{Proceedings of the European Conference on Computer Vision
  (ECCV)}.

\bibitem[{Berthelot et~al.(2019)Berthelot, Carlini, Goodfellow, Papernot,
  Oliver, and Raffel}]{MixMatch}
Berthelot, D.; Carlini, N.; Goodfellow, I.; Papernot, N.; Oliver, A.; and
  Raffel, C.~A. 2019.
\newblock MixMatch: A Holistic Approach to Semi-Supervised Learning.
\newblock In \emph{Advances in Neural Information Processing Systems}.

\bibitem[{Bertinetto et~al.(2019)Bertinetto, Henriques, Torr, and
  Vedaldi}]{bertinetto2018metalearning}
Bertinetto, L.; Henriques, J.~F.; Torr, P.; and Vedaldi, A. 2019.
\newblock Meta-learning with differentiable closed-form solvers.
\newblock In \emph{International Conference on Learning Representations}.

\bibitem[{Boudiaf et~al.(2020)Boudiaf, Ziko, Rony, Dolz, Piantanida, and
  Ben~Ayed}]{NEURIPS2020_196f5641}
Boudiaf, M.; Ziko, I.; Rony, J.; Dolz, J.; Piantanida, P.; and Ben~Ayed, I.
  2020.
\newblock Information Maximization for Few-Shot Learning.
\newblock In \emph{Advances in Neural Information Processing Systems}.

\bibitem[{Cao, Brbic, and Leskovec(2021)}]{cao2021concept}
Cao, K.; Brbic, M.; and Leskovec, J. 2021.
\newblock Concept Learners for Few-Shot Learning.
\newblock In \emph{International Conference on Learning Representations}.

\bibitem[{Chen et~al.(2020)Chen, Kornblith, Swersky, Norouzi, and
  Hinton}]{simclrv2}
Chen, T.; Kornblith, S.; Swersky, K.; Norouzi, M.; and Hinton, G.~E. 2020.
\newblock Big Self-Supervised Models are Strong Semi-Supervised Learners.
\newblock In \emph{Advances in Neural Information Processing Systems}.

\bibitem[{Chen et~al.(2019)Chen, Liu, Kira, Wang, and Huang}]{chen2018a}
Chen, W.-Y.; Liu, Y.-C.; Kira, Z.; Wang, Y.-C.~F.; and Huang, J.-B. 2019.
\newblock A Closer Look at Few-shot Classification.
\newblock In \emph{International Conference on Learning Representations}.

\bibitem[{Chen, Maji, and Learned-Miller(2021)}]{Chen_2021_CVPR}
Chen, Z.; Maji, S.; and Learned-Miller, E. 2021.
\newblock Shot in the Dark: Few-Shot Learning With No Base-Class Labels.
\newblock In \emph{Proceedings of the IEEE/CVF Conference on Computer Vision
  and Pattern Recognition (CVPR) Workshops}, 2668--2677.

\bibitem[{{Deng} et~al.(2009){Deng}, {Dong}, {Socher}, {Li}, {Kai Li}, and {Li
  Fei-Fei}}]{ImageNet}
{Deng}, J.; {Dong}, W.; {Socher}, R.; {Li}, L.; {Kai Li}; and {Li Fei-Fei}.
  2009.
\newblock ImageNet: A large-scale hierarchical image database.
\newblock In \emph{2009 IEEE Conference on Computer Vision and Pattern
  Recognition}.

\bibitem[{Devlin et~al.(2019)Devlin, Chang, Lee, and
  Toutanova}]{devlin-etal-2019-bert}
Devlin, J.; Chang, M.-W.; Lee, K.; and Toutanova, K. 2019.
\newblock {BERT}: Pre-training of Deep Bidirectional Transformers for Language
  Understanding.
\newblock In \emph{Proceedings of the 2019 Conference of the North {A}merican
  Chapter of the Association for Computational Linguistics}, 4171--4186.
  Association for Computational Linguistics.

\bibitem[{Dhillon et~al.(2020)Dhillon, Chaudhari, Ravichandran, and
  Soatto}]{Dhillon2020A}
Dhillon, G.~S.; Chaudhari, P.; Ravichandran, A.; and Soatto, S. 2020.
\newblock A Baseline for Few-Shot Image Classification.
\newblock In \emph{International Conference on Learning Representations}.

\bibitem[{Donahue et~al.(2014)Donahue, Jia, Vinyals, Hoffman, Zhang, Tzeng, and
  Darrell}]{pmlr-v32-donahue14}
Donahue, J.; Jia, Y.; Vinyals, O.; Hoffman, J.; Zhang, N.; Tzeng, E.; and
  Darrell, T. 2014.
\newblock DeCAF: A Deep Convolutional Activation Feature for Generic Visual
  Recognition.
\newblock In \emph{Proceedings of the 31st International Conference on Machine
  Learning}, 647--655. PMLR.

\bibitem[{Esfandiarpoor, Hajabdollahi, and Bach(2020)}]{esfandiarpoor:arxiv20}
Esfandiarpoor, R.; Hajabdollahi, M.; and Bach, S.~H. 2020.
\newblock Pseudo Shots: {F}ew-Shot Learning with Auxiliary Data.
\newblock \emph{arXiv preprint arXiv:2012.07176}.

\bibitem[{Fei et~al.(2021)Fei, Lu, Xiang, and Huang}]{fei2021melr}
Fei, N.; Lu, Z.; Xiang, T.; and Huang, S. 2021.
\newblock {\{}MELR{\}}: Meta-Learning via Modeling Episode-Level Relationships
  for Few-Shot Learning.
\newblock In \emph{International Conference on Learning Representations}.

\bibitem[{Finn, Abbeel, and Levine(2017)}]{MAML}
Finn, C.; Abbeel, P.; and Levine, S. 2017.
\newblock Model-Agnostic Meta-Learning for Fast Adaptation of Deep Networks.
\newblock In \emph{Proceedings of the 34th International Conference on Machine
  Learning}, 1126--1135.

\bibitem[{Flennerhag et~al.(2020)Flennerhag, Rusu, Pascanu, Visin, Yin, and
  Hadsell}]{Flennerhag2020Meta-Learning}
Flennerhag, S.; Rusu, A.~A.; Pascanu, R.; Visin, F.; Yin, H.; and Hadsell, R.
  2020.
\newblock Meta-Learning with Warped Gradient Descent.
\newblock In \emph{International Conference on Learning Representations}.

\bibitem[{Furlanello et~al.(2018)Furlanello, Lipton, Tschannen, Itti, and
  Anandkumar}]{pmlr-v80-furlanello18a}
Furlanello, T.; Lipton, Z.; Tschannen, M.; Itti, L.; and Anandkumar, A. 2018.
\newblock Born Again Neural Networks.
\newblock In \emph{Proceedings of the 35th International Conference on Machine
  Learning}, Proceedings of Machine Learning Research, 1607--1616. PMLR.

\bibitem[{Gao et~al.(2021)Gao, Fei, Liu, Lu, Xiang, and
  Huang}]{gao2021contrastive}
Gao, Y.; Fei, N.; Liu, G.; Lu, Z.; Xiang, T.; and Huang, S. 2021.
\newblock Contrastive Prototype Learning with Augmented Embeddings for Few-Shot
  Learning.
\newblock \emph{arXiv preprint arXiv:2101.09499}.

\bibitem[{He et~al.(2016)He, Zhang, Ren, and Sun}]{He_2016_CVPR}
He, K.; Zhang, X.; Ren, S.; and Sun, J. 2016.
\newblock Deep Residual Learning for Image Recognition.
\newblock In \emph{Proceedings of the IEEE Conference on Computer Vision and
  Pattern Recognition (CVPR)}.

\bibitem[{Hinton, Vinyals, and Dean(2015)}]{KD}
Hinton, G.; Vinyals, O.; and Dean, J. 2015.
\newblock Distilling the Knowledge in a Neural Network.
\newblock In \emph{NIPS Deep Learning and Representation Learning Workshop}.

\bibitem[{Huang et~al.(2021)Huang, Wu, Li, Huo, and Gao}]{HUANG2021107935}
Huang, H.; Wu, Z.; Li, W.; Huo, J.; and Gao, Y. 2021.
\newblock Local descriptor-based multi-prototype network for few-shot Learning.
\newblock \emph{Pattern Recognition}, 116: 107935.

\bibitem[{Koch(2015)}]{Koch2015SiameseNN}
Koch, G.~R. 2015.
\newblock Siamese Neural Networks for One-Shot Image Recognition.
\newblock In \emph{ICML deep learning workshop}.

\bibitem[{Krizhevsky, Sutskever, and Hinton(2012)}]{AlexNet}
Krizhevsky, A.; Sutskever, I.; and Hinton, G.~E. 2012.
\newblock ImageNet Classification with Deep Convolutional Neural Networks.
\newblock In \emph{Advances in Neural Information Processing Systems}.

\bibitem[{Lazarou, Avrithis, and Stathaki(2020)}]{lazarou2020iterative}
Lazarou, M.; Avrithis, Y.; and Stathaki, T. 2020.
\newblock Iterative label cleaning for transductive and semi-supervised
  few-shot learning.
\newblock \emph{arXiv preprint arXiv:2012.07962}.

\bibitem[{Lee(2013)}]{pseudo-label}
Lee, D.-H. 2013.
\newblock Pseudo-Label : The Simple and Efficient Semi-Supervised Learning
  Method for Deep Neural Networks.
\newblock \emph{ICML 2013 Workshop : Challenges in Representation Learning
  (WREPL)}.

\bibitem[{Lee et~al.(2019)Lee, Maji, Ravichandran, and Soatto}]{Lee_2019_CVPR}
Lee, K.; Maji, S.; Ravichandran, A.; and Soatto, S. 2019.
\newblock Meta-Learning With Differentiable Convex Optimization.
\newblock In \emph{Proceedings of the IEEE/CVF Conference on Computer Vision
  and Pattern Recognition (CVPR)}.

\bibitem[{Li et~al.(2019{\natexlab{a}})Li, Wang, Xu, Huo, Yang, and
  Luo}]{li2019DN4}
Li, W.; Wang, L.; Xu, J.; Huo, J.; Yang, G.; and Luo, J. 2019{\natexlab{a}}.
\newblock Revisiting Local Descriptor based Image-to-Class Measure for Few-shot
  Learning.
\newblock In \emph{CVPR}.

\bibitem[{Li et~al.(2019{\natexlab{b}})Li, Sun, Liu, Zhou, Zheng, Chua, and
  Schiele}]{NEURIPS2019_bf25356f}
Li, X.; Sun, Q.; Liu, Y.; Zhou, Q.; Zheng, S.; Chua, T.-S.; and Schiele, B.
  2019{\natexlab{b}}.
\newblock Learning to Self-Train for Semi-Supervised Few-Shot Classification.
\newblock In \emph{Advances in Neural Information Processing Systems}.

\bibitem[{Li et~al.(2017)Li, Zhou, Chen, and Li}]{li2017metasgd}
Li, Z.; Zhou, F.; Chen, F.; and Li, H. 2017.
\newblock Meta-SGD: Learning to Learn Quickly for Few-Shot Learning.
\newblock \emph{arXiv preprint arXiv:1707.09835}.

\bibitem[{Lichtenstein et~al.(2020)Lichtenstein, Sattigeri, Feris, Giryes, and
  Karlinsky}]{TAFSSL}
Lichtenstein, M.; Sattigeri, P.; Feris, R.; Giryes, R.; and Karlinsky, L. 2020.
\newblock TAFSSL: Task-Adaptive Feature Sub-Space Learning for Few-Shot
  Classification.
\newblock In Vedaldi, A.; Bischof, H.; Brox, T.; and Frahm, J.-M., eds.,
  \emph{Computer Vision -- ECCV 2020}.

\bibitem[{Liu et~al.(2020)Liu, Cao, Lin, Li, Zhang, Long, and Hu}]{neg-cosine}
Liu, B.; Cao, Y.; Lin, Y.; Li, Q.; Zhang, Z.; Long, M.; and Hu, H. 2020.
\newblock Negative Margin Matters: Understanding Margin in Few-Shot
  Classification.
\newblock In Vedaldi, A.; Bischof, H.; Brox, T.; and Frahm, J.-M., eds.,
  \emph{Computer Vision -- ECCV 2020}.

\bibitem[{Liu et~al.(2019)Liu, Lee, Park, Kim, Yang, Hwang, and
  Yang}]{liu2018learning}
Liu, Y.; Lee, J.; Park, M.; Kim, S.; Yang, E.; Hwang, S.; and Yang, Y. 2019.
\newblock Learning to Propagate Labels: Transductive Propagation Network for
  Few-shot Learning.
\newblock In \emph{International Conference on Learning Representations}.

\bibitem[{Oreshkin, López, and Lacoste(2018)}]{TADAM}
Oreshkin, B.~N.; López, P.~R.; and Lacoste, A. 2018.
\newblock TADAM: Task dependent adaptive metric for improved few-shot learning.
\newblock In \emph{NeurIPS}.

\bibitem[{Pham et~al.(2021)Pham, Dai, Xie, Luong, and Le}]{pham2021meta}
Pham, H.; Dai, Z.; Xie, Q.; Luong, M.-T.; and Le, Q.~V. 2021.
\newblock Meta Pseudo Labels.
\newblock \emph{arXiv preprint arXiv:2003.10580}.

\bibitem[{Rajasegaran et~al.(2020)Rajasegaran, Khan, Hayat, Khan, and
  Shah}]{rajasegaran2020selfsupervised}
Rajasegaran, J.; Khan, S.; Hayat, M.; Khan, F.~S.; and Shah, M. 2020.
\newblock Self-supervised Knowledge Distillation for Few-shot Learning.
\newblock \emph{arXiv preprint arXiv:2006.09785}.

\bibitem[{Ravi and Larochelle(2017)}]{Ravi2017OptimizationAA}
Ravi, S.; and Larochelle, H. 2017.
\newblock Optimization as a Model for Few-Shot Learning.
\newblock In \emph{ICLR}.

\bibitem[{Ravichandran, Bhotika, and Soatto(2019)}]{Ravichandran_2019_ICCV}
Ravichandran, A.; Bhotika, R.; and Soatto, S. 2019.
\newblock Few-Shot Learning With Embedded Class Models and Shot-Free Meta
  Training.
\newblock In \emph{Proceedings of the IEEE/CVF International Conference on
  Computer Vision (ICCV)}.

\bibitem[{Ren et~al.(2018)Ren, Ravi, Triantafillou, Snell, Swersky, Tenenbaum,
  Larochelle, and Zemel}]{ren2018metalearning}
Ren, M.; Ravi, S.; Triantafillou, E.; Snell, J.; Swersky, K.; Tenenbaum, J.~B.;
  Larochelle, H.; and Zemel, R.~S. 2018.
\newblock Meta-Learning for Semi-Supervised Few-Shot Classification.
\newblock In \emph{International Conference on Learning Representations}.

\bibitem[{Rizve et~al.(2021)Rizve, Khan, Khan, and Shah}]{rizve2021exploring}
Rizve, M.~N.; Khan, S.; Khan, F.~S.; and Shah, M. 2021.
\newblock Exploring Complementary Strengths of Invariant and Equivariant
  Representations for Few-Shot Learning.
\newblock \emph{arXiv preprint arXiv:2103.01315}.

\bibitem[{Rusu et~al.(2019)Rusu, Rao, Sygnowski, Vinyals, Pascanu, Osindero,
  and Hadsell}]{rusu2018metalearning}
Rusu, A.~A.; Rao, D.; Sygnowski, J.; Vinyals, O.; Pascanu, R.; Osindero, S.;
  and Hadsell, R. 2019.
\newblock Meta-Learning with Latent Embedding Optimization.
\newblock In \emph{International Conference on Learning Representations}.

\bibitem[{Senior et~al.(2020)Senior, Evans, Jumper, Kirkpatrick, Sifre, Green,
  Qin, Žídek, Nelson, Bridgland, Penedones, Petersen, Simonyan, Crossan,
  Kohli, Jones, Silver, Kavukcuoglu, and Hassabis}]{AlphaFold}
Senior, A.; Evans, R.; Jumper, J.; Kirkpatrick, J.; Sifre, L.; Green, T.; Qin,
  C.; Žídek, A.; Nelson, A.; Bridgland, A.; Penedones, H.; Petersen, S.;
  Simonyan, K.; Crossan, S.; Kohli, P.; Jones, D.; Silver, D.; Kavukcuoglu, K.;
  and Hassabis, D. 2020.
\newblock Improved protein structure prediction using potentials from deep
  learning.
\newblock \emph{Nature}, 577: 1--5.

\bibitem[{Shen et~al.(2021)Shen, Liu, Qin, Savvides, and
  Cheng}]{DBLP:journals/corr/abs-2102-03983}
Shen, Z.; Liu, Z.; Qin, J.; Savvides, M.; and Cheng, K. 2021.
\newblock Partial Is Better Than All: Revisiting Fine-tuning Strategy for
  Few-shot Learning.
\newblock \emph{CoRR}, abs/2102.03983.

\bibitem[{Simon et~al.(2020)Simon, Koniusz, Nock, and
  Harandi}]{Simon_2020_CVPR}
Simon, C.; Koniusz, P.; Nock, R.; and Harandi, M. 2020.
\newblock Adaptive Subspaces for Few-Shot Learning.
\newblock In \emph{Proceedings of the IEEE/CVF Conference on Computer Vision
  and Pattern Recognition (CVPR)}.

\bibitem[{Snell, Swersky, and Zemel(2017)}]{ProtoNet}
Snell, J.; Swersky, K.; and Zemel, R. 2017.
\newblock Prototypical Networks for Few-shot Learning.
\newblock In \emph{Advances in Neural Information Processing Systems}.

\bibitem[{Sohn et~al.(2020)Sohn, Berthelot, Carlini, Zhang, Zhang, Raffel,
  Cubuk, Kurakin, and Li}]{fixmatch}
Sohn, K.; Berthelot, D.; Carlini, N.; Zhang, Z.; Zhang, H.; Raffel, C.~A.;
  Cubuk, E.~D.; Kurakin, A.; and Li, C.-L. 2020.
\newblock FixMatch: Simplifying Semi-Supervised Learning with Consistency and
  Confidence.
\newblock In \emph{Advances in Neural Information Processing Systems},
  596--608.

\bibitem[{Sun et~al.(2019)Sun, Liu, Chua, and Schiele}]{Sun_2019_CVPR}
Sun, Q.; Liu, Y.; Chua, T.-S.; and Schiele, B. 2019.
\newblock Meta-Transfer Learning for Few-Shot Learning.
\newblock In \emph{Proceedings of the IEEE/CVF Conference on Computer Vision
  and Pattern Recognition (CVPR)}.

\bibitem[{Sung et~al.(2018)Sung, Yang, Zhang, Xiang, Torr, and
  Hospedales}]{Sung_2018_CVPR}
Sung, F.; Yang, Y.; Zhang, L.; Xiang, T.; Torr, P.~H.; and Hospedales, T.~M.
  2018.
\newblock Learning to Compare: Relation Network for Few-Shot Learning.
\newblock In \emph{Proceedings of the IEEE Conference on Computer Vision and
  Pattern Recognition (CVPR)}.

\bibitem[{Tian et~al.(2020)Tian, Wang, Krishnan, Tenenbaum, and
  Isola}]{tian2020rethink}
Tian, Y.; Wang, Y.; Krishnan, D.; Tenenbaum, J.~B.; and Isola, P. 2020.
\newblock Rethinking Few-Shot Image Classification: A Good Embedding is All You
  Need?
\newblock In Vedaldi, A.; Bischof, H.; Brox, T.; and Frahm, J.-M., eds.,
  \emph{Computer Vision -- ECCV 2020}.

\bibitem[{Vinyals et~al.(2016)Vinyals, Blundell, Lillicrap, kavukcuoglu, and
  Wierstra}]{MatchNet}
Vinyals, O.; Blundell, C.; Lillicrap, T.; kavukcuoglu, k.; and Wierstra, D.
  2016.
\newblock Matching Networks for One Shot Learning.
\newblock In \emph{Advances in Neural Information Processing Systems}.

\bibitem[{Wang et~al.(2020)Wang, Xu, Liu, Zhang, and Fu}]{Wang_2020_CVPR}
Wang, Y.; Xu, C.; Liu, C.; Zhang, L.; and Fu, Y. 2020.
\newblock Instance Credibility Inference for Few-Shot Learning.
\newblock In \emph{Proceedings of the IEEE/CVF Conference on Computer Vision
  and Pattern Recognition (CVPR)}.

\bibitem[{Wei et~al.(2021)Wei, Shen, Chen, and Ma}]{wei2021theoretical}
Wei, C.; Shen, K.; Chen, Y.; and Ma, T. 2021.
\newblock Theoretical Analysis of Self-Training with Deep Networks on Unlabeled
  Data.
\newblock In \emph{International Conference on Learning Representations}.

\bibitem[{Xu et~al.(2021)Xu, yifan xu, Wang, and Tu}]{xu2021attentional}
Xu, W.; yifan xu; Wang, H.; and Tu, Z. 2021.
\newblock Attentional Constellation Nets for Few-Shot Learning.
\newblock In \emph{International Conference on Learning Representations}.

\bibitem[{Ye et~al.(2020)Ye, Hu, Zhan, and Sha}]{Ye_2020_CVPR}
Ye, H.-J.; Hu, H.; Zhan, D.-C.; and Sha, F. 2020.
\newblock Few-Shot Learning via Embedding Adaptation With Set-to-Set Functions.
\newblock In \emph{Proceedings of the IEEE/CVF Conference on Computer Vision
  and Pattern Recognition (CVPR)}.

\bibitem[{Yoon, Seo, and Moon(2019)}]{pmlr-v97-yoon19a}
Yoon, S.~W.; Seo, J.; and Moon, J. 2019.
\newblock {T}ap{N}et: Neural Network Augmented with Task-Adaptive Projection
  for Few-Shot Learning.
\newblock In \emph{Proceedings of the 36th International Conference on Machine
  Learning}, Proceedings of Machine Learning Research, 7115--7123. PMLR.

\bibitem[{Yu et~al.(2020)Yu, Chen, Cheng, and Luo}]{Yu_2020_CVPR}
Yu, Z.; Chen, L.; Cheng, Z.; and Luo, J. 2020.
\newblock TransMatch: A Transfer-Learning Scheme for Semi-Supervised Few-Shot
  Learning.
\newblock In \emph{Proceedings of the IEEE/CVF Conference on Computer Vision
  and Pattern Recognition (CVPR)}.

\bibitem[{Zhang et~al.(2020)Zhang, Cai, Lin, and Shen}]{Zhang_2020_CVPR}
Zhang, C.; Cai, Y.; Lin, G.; and Shen, C. 2020.
\newblock DeepEMD: Few-Shot Image Classification With Differentiable Earth
  Mover's Distance and Structured Classifiers.
\newblock In \emph{IEEE/CVF Conference on Computer Vision and Pattern
  Recognition (CVPR)}.

\bibitem[{Zhang et~al.(2021)Zhang, Zhang, Lu, Xiang, Ding, and
  Huang}]{zhang2021iept}
Zhang, M.; Zhang, J.; Lu, Z.; Xiang, T.; Ding, M.; and Huang, S. 2021.
\newblock {\{}IEPT{\}}: Instance-Level and Episode-Level Pretext Tasks for
  Few-Shot Learning.
\newblock In \emph{International Conference on Learning Representations}.

\bibitem[{Zhou et~al.(2020)Zhou, Cui, Jia, Yang, and Tian}]{Zhou_2020_CVPR}
Zhou, L.; Cui, P.; Jia, X.; Yang, S.; and Tian, Q. 2020.
\newblock Learning to Select Base Classes for Few-Shot Classification.
\newblock In \emph{Proceedings of the IEEE/CVF Conference on Computer Vision
  and Pattern Recognition (CVPR)}.

\bibitem[{Zhu et~al.(2021)Zhu, Li, Liao, and Luo}]{ZHU2021107797}
Zhu, W.; Li, W.; Liao, H.; and Luo, J. 2021.
\newblock Temperature network for few-shot learning with distribution-aware
  large-margin metric.
\newblock \emph{Pattern Recognition}, 112: 107797.

\end{thebibliography}
}

\end{document}